\documentclass{article}

\usepackage{PRIMEarxiv}

\usepackage[utf8]{inputenc} % allow utf-8 input
\usepackage[T1]{fontenc}    % use 8-bit T1 fonts
\usepackage{hyperref}       % hyperlinks
\usepackage{url}            % simple URL typesetting
\usepackage{booktabs}       % professional-quality tables
\usepackage{nicefrac}       % compact symbols for 1/2, etc.
\usepackage{microtype}      % microtypography
\usepackage{lipsum}
\usepackage{fancyhdr}       % header
\usepackage{graphicx}       % graphics
\usepackage{amsmath,amsfonts}
\usepackage{subfigure,multirow,array,pbox,epstopdf}

\title{On Characterizing the Evolution of Embedding Space of Neural Networks using Algebraic Topology
}

\author{ 
	Suryaka suresh \\
 Infosys Centre for AI \\ Department of Computer Science \\ IIIT Delhi, India \\
 suryakas@iiitd.ac.in \\
	\And
	Vinayak Abrol \\
   Infosys Centre for AI \\ Department of Computer Science \\ IIIT Delhi, India \\
   abrol@iiitd.ac.in \\
	\AND
	\hspace{1mm} Bishoy Das \\
   Department of Electrical Engineering \\ IIT Delhi India \\
   bishshoy.das@ee.iitd.ac.in \\
	\And
	\hspace{1mm} Sumantra Dutta Roy \\
   Department of Electrical Engineering \\ IIT Delhi India \\
  sumantra@ee.iitd.ac.in \\
}

\begin{document}
\maketitle

\begin{abstract}
We study how the topology of feature embedding space changes as it passes through the layers of a well-trained deep neural network (DNN) through Betti numbers. Motivated by existing studies using simplicial complexes on shallow fully connected networks (FCN), we present an extended analysis using Cubical homology instead, with a variety of popular deep architectures and real image datasets. We demonstrate that as depth increases, a topologically complicated dataset is transformed into a simple one, resulting in Betti numbers attaining their lowest possible value. The rate of decay in topological complexity (as a metric) helps quantify the impact of architectural choices on the generalization ability. Interestingly from a representation learning perspective, we highlight several invariances such as topological invariance of (1) an architecture on similar datasets; (2)  embedding space of a dataset for architectures of variable depth; (3) embedding space to input resolution/size, and (4) data sub-sampling. In order to further demonstrate the link between expressivity \& the generalization capability of a network, we consider the task of ranking pre-trained models for downstream classification task (transfer learning). Compared to existing approaches, the proposed metric has a better correlation to the actually achievable accuracy via fine-tuning the pre-trained model. 
\end{abstract}


\section{Introduction}
\label{sec:intro}
Recently, deep learning has attracted a lot of attention in various learning tasks, as they offer great representational power as a result of non-linear transformation based layered architectures \cite{Guo21,wei22}. A deep neural network's (DNN's) architectural components consist of affine-linear maps and nonlinearities, a combination of which is often termed as ``black boxes” that lack explainability. While many properties such as expressivity, generalization error, sample/model complexity, and loss landscape of the whole network have been studied extensively in various mathematical settings \cite{c:8}, the study of the evolution of feature embedding space has received little attention. Most existing studies linking expressivity and generalization in DNNs \cite{c:8,c:6,MURRAY22} emphasize the effect of depth on the learning dynamics in terms of Jacobian or Hessian of the network. However, such studies don't answer the question of which model is the most compatible for a dataset and task so as to provide practitioners valuable insights for empirical deep learning.

\begin{figure*}[ht]
    \centering
    \includegraphics[width=.85\textwidth]{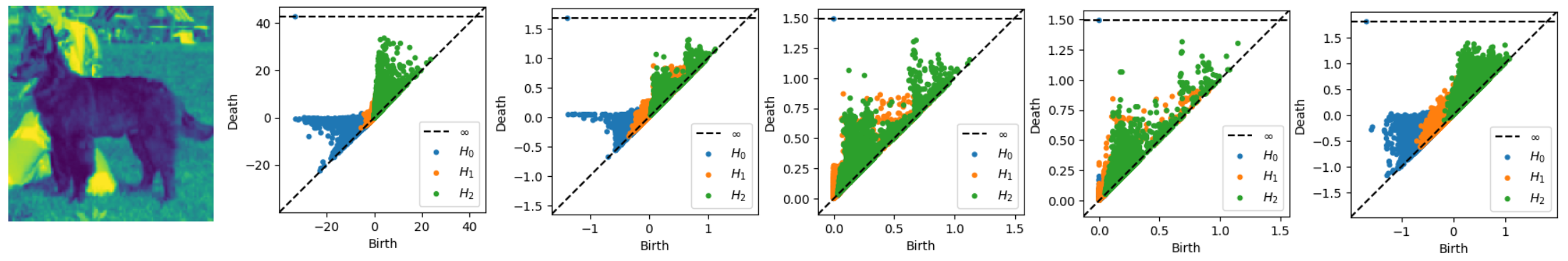}
    \includegraphics[width=.85\textwidth]{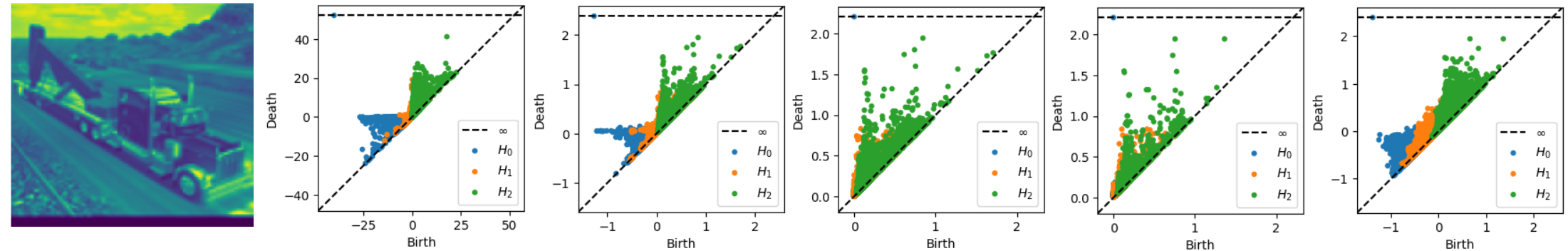}
    \caption{Persistence diagrams of embedding space of (from the left) Conv2d, BatchNorm2d, ReLU, MaxPool2d, and denseblock1, of DenseNet-169 for the images of the STL-10 dataset, respectively. $H_{i}$ in the persistence diagram is the $i^{th}-$homology and homology is the algebraic object that defines the topology of a given space, here RGB image. Note how topological features at the block level are more interpretable compared to ones at individual layers like ReLU and MaxPool2d.}
    \label{fig1}
\end{figure*}

In this work, we aim to analyze the layer-by-layer embedding space of a DNN and leverage them to understand their learning characteristic empirically. Specifically, we use Algebraic topology \cite{c:13} to capture high-order dependencies of embedding space in a data-dependent setting with popular pre-trained networks. Recently, topological tools such as Persistent Homology \cite{c:13}, and K-mapper \cite{c:10} have been widely explored to study DNNs, though restricted to shallow networks or low dimensional problems. We study the expressivity of a DNN in terms of extracted Betti numbers from the persistence diagram of the embedding space. Persistence homology maps complex data to persistence diagrams, a simple persistence-based feature vectorization that visualizes a concise summary of topological features in data. For statistical analysis, the persistence diagram needs to be equipped with a function that maps to this vector space like Betti numbers which is a quantification of the multiplicity of the homology. For a point cloud, Betti numbers represent a count of the topology features that appears in the subcomplex during filtration. The computational complexity involved in conventional algebraic topology tools is very high \cite{c:6}, and hence we present our analysis using Cubical homology \cite{c:3} based on the method of cubical filtration that has been shown to be more efficient in extracting topological features of higher dimensional data like images. As an illustration, Fig. \ref{fig1} shows the persistent diagrams corresponding to two example images from the STL-10 dataset of the embedding space of a DenseNet-169 model (pre-trained on ImageNet). One can attribute the similarities in the persistent diagrams to the network's learning characteristics (feature extraction) from images while the dissimilarities to the image-specific attributes. Thus this provides a model-agnostic way of characterizing the embedding space via the quantification of persistent diagrams using Betti numbers. Experiments with a variety of popular deep architectures and real image datasets demonstrate that as depth increases, a topologically complicated dataset is transformed into a simple one, resulting in Betti numbers attaining their lowest possible value. Thus, we exploit the rate of decay in topological complexity $\Omega$ \cite{c:9,c:4} (as a metric), an alternative to the popular Euler characteristic, to quantify the impact of architectural choices on the generalization ability. While doing so, we highlight and establish several interesting invariances such as the topological invariance of: 

\noindent\textit{-an architecture on similar datasets.}

\noindent\textit{-embedding of a dataset for an architecture of variable depth.}

\noindent\textit{-embedding space to input resolution/size.}

\noindent\textit{-embedding space to data sub-sampling.}

\noindent We extend the work of Naitzat et al. (2020) \cite{c:4} to multi-class higher dimensional problem settings. We leverage our analysis to establish a link between expressivity and the generalization capability of a network by considering the task of ranking pre-trained models for downstream supervised classification aka transfer learning. As explained earlier, we use the rate of decay of Betti numbers across layers as a metric via curve fitting and demonstrated an inversely proportional relationship between the slope and the actually achievable accuracy of a fine-tuned pre-trained network.

\section{Related work} 
Carlsson et al. (2008) \cite{c:20} and follow-up works from the group demonstrated how topological features serve as a potential measure of generalization in DNNs. However, their study was mainly restricted to visualization of the weight space of shallow networks (in terms of manifolds with well-known topology e.g., Klein bottle) trained on toy datasets such as MNIST using the Mapper algorithm \cite{c:10}. Similarly, Rieck et al (2019)~\cite{c:17} used topology to track the evolution of DNN weights during training and proposed an early stopping criterion. In contrast, a series of works have attempted to study how DNNs transform a complex dataset to a topologically simpler one using Betti numbers. Bianchini et al. (2014) \cite{c:16} proposed a theoretical measure to evaluate the complexity of DNN function space. Guss et al. (2018) \cite{c:6} uses the Betti number as a measure of data complexity to characterize neural networks by the depth required to resolve the complexity. Hamada et al. (2018) \cite{c:18} presented a data-driven test to establish if the transformed dataset is topologically simple. It doesn't provide a quantification, rather just confirms if $0^{th}$ order Betti number is greater than 1. Naitzat et al. (2020) \cite{c:4} explored persistent homology of the dataset as it passes through layers of a DNN. However, the study is restricted to shallow feed-forward networks and simulated dataset (binary labels) to prove the argument since the Betti number of a dataset are unknown in practice. Akai et al. (2021) \cite{c:19} extended the work of \cite{c:4} to study the stability of DNN mapping and the impact of outliers, but their analysis doesn't hold in high dimensional settings.

In contrast, work in this paper focuses on the representational power of a network with respect to a task (and dataset) at hand. We demonstrate the generalizability of a model through the evolution of the topological complexity of layer-by-layer embedding space. Compared to existing studies, the computational complexity around scaling a topological method to higher dimensional multi-class problems is tackled using recent advancements in Cubical homology. Finally, experiments in this work are done using a variety of popular architectures and real-world datasets.  

\section{Topological Preliminaries}
In this section, we describe the relevant topological definitions based on the mathematical setting in \cite{c:1}. 

\begin{figure}[ht]
    \centering
    \includegraphics[width=0.25\textwidth]{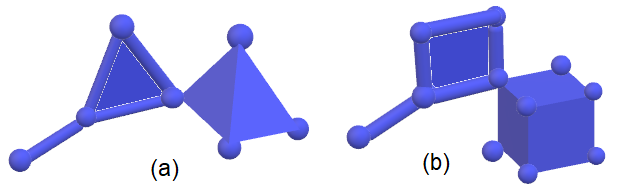}
    \caption{Illustration of (a) simplicial complex and (b) cubical complex. The spheres represent the vertex i.e., a 0-dimensional simplex. Similarly, $1$-simplex is an edge, $2$-simplex is a triangle, and $3$-simplex is a tetrahedron. %For an image, weighted cubical complex consisting of cubical cells, 0-dimensional is a pixel/voxel
    }
    \label{fig2}
\end{figure}

\subsection{Persistence homology}
Given a space $X$ defined in terms of a point cloud $\in\mathbb{R}^d$ with $N$ points , the topological features can be computed using persistence homology. It requires the space to be represented using a simplicial complex i.e., constructing open sets around each data point using their intersection properties. This simplicial complex defines properties such as the number of connected components, holes, and voids it contains. Using a grid of scale values, a sequence of simplicial complexes is obtained to get information about the most persistent features that likely represent true underlying features of the shape of data. In order to work with image datasets (voxel data), recently cubical complexes have been shown to be efficient due to their robustness in higher dimensional problems \cite{c:12}. These are equivalent to simplicial complexes with basic components such as squares, cubes, and higher dimensional analogs (see Fig. \ref{fig2}). The topological information is often summarized as a persistence diagram (see Fig. \ref{fig1}); a collection of points $(x,y)\in\mathbb{R}^2$ representing a topological feature born at scale $x$ and persisting until scale $y$. Points at infinity i.e. homology that never dies, are represented via an infinity bar in the diagram. Note: $\beta_{i}$ is the $H_{i}$ that exists at a threshold value.

\subsection{Quantification of Topology using Betti numbers}
Using a persistent diagram, a topology can be quantified in terms of readily commutable Betti numbers $\beta_k(X)$ that are the simplest invariants representing the shape of a manifold. For $X\in\mathbb{R}^d$, there are at most $d$ non-zero Betti numbers. Typically, due to computational complexity constraints, the first three i.e., $0^{th}$ Betti number $\beta_{0}(X)$, $\beta_{1}(X)$ and $\beta_{2}(X)$ that counts the connected components, holes, and voids are used. Readers are encouraged to refer to Section 2 from Naitzat et al. (2020) \cite{c:4} for more details and examples. Here, the Betti curve is a function that maps the persistence diagram to an integer value over a range of threshold values $\eta$. Then $\beta_k$ are computed for an appropriate value of $\eta$. We consider the embedding space of an image as the geometric object of interest, which is molded through homeomorphic or non-homeomorphic maps based on the task at hand. For instance, segmentation of an image would require a homeomorphic map in order to preserve the topology \cite{c:22} while classification requires a non-homeomorphic map to collapse the space of images (each class) to a single point \cite{c:4}. This work mainly focuses on the classification task and hence uses the Betti numbers, which is a unique signature for an object (here an image). Following \cite{c:3} we use the voxel embedding space of an image $x$ to compute $\beta_k(x)$ which are combined to obtain a measure as:

\begin{eqnarray}
\label{eq1}
 &\Omega = \displaystyle \frac{1}{N}\sum_{i=1}^N \omega_{i} \\ \nonumber
 &\omega_i = \beta_{0}(x_i)+\beta_{1}(x_i) + \ldots +\beta_{d}(x_i); \;\; x_i \in X 
\end{eqnarray}
Here, $\omega$ is the topological complexity of the embedding space of an image and  $\Omega$ is the expected topological complexity of the embedding space for a dataset. Note that eq(\ref{eq1})  is a different measure than the one used in classical topology appearing in Morse theory \cite{c:21} and adapted in \cite{c:9,c:4}. Later we empirically demonstrate that this measure is robust to data sampling as well as input resolution to a reasonable extent.

\section{Experimental Setup}
\label{sec:exp}
In this section, we provide details of the experimental protocol and software setup we used to carry out our experiments. We also delineate the different neural network architectures and datasets used in our study. All the experiments can be reproduced using the implementation available on GitHub\footnote{\url{https://github.com/Cross-Caps/DNNTopology}}.

\subsection{Datasets}
All the main experiments are done using a test set of three datasets (unless stated otherwise): STL-10 \cite{coates2011stl10}, CIFAR-10 \cite{Krizhevsky09learningmultiple} and Aircraft \cite{maji13fine-grained}. The STL-10 dataset consists of 500 training and 800 test images of $96\times96$ resolution dispersed across 10 classes. The CIFAR-10 dataset consists of 50000 training and 10000 test images spread across 10 classes, with each image having a resolution of $32\times32$. Both STL-10 and CIFAR-10 pertain to the use cases where the target number of classes is small. Hence, we have also the test set of the Aircraft dataset consisting of 10200 images divided into 102 classes.

\subsection{Neural Network Architectures}
\begin{enumerate}
    \item {VGG} \cite{simonyan2014very} is a sequential feedforward neural network with a uniform collection of convolutional layers followed by ReLU (Rectified Linear Unit) nonlinearity.
    \item {ResNet} (or Residual Networks) \cite{he2016deep} is a class of networks that features a skip connection that controls how much information flows to the next layer.
    \item {DenseNet} \cite{huang2017densely} have transition layers that connect the dense blocks which spreads their weight function all over the neural network and allows feature reuse.
    \item {MobileNetV2} \cite{sandler2018mobilenetv2} is a lightweight neural network that uses depthwise separable convolutions. The residual blocks have a $wide\rightarrow narrow \rightarrow wide$ approach.
\end{enumerate}
\noindent In total, we have experimented with 9 networks namely: VGG-16, ResNet-18, ResNet-50, ResNet-101, ResNet-152, MobileNetV2, DenseNet-121, DenseNet-169 and DenseNet-201. These models obtained from PyTorch model zoo are pre-trained on ImageNet dataset \cite{ILSVRC15}. They use an input image size of $256\times256$, center cropped to $224\times224$, and coupled with standard data augmentations.

\subsection{Computation of Betti numbers}
This work uses Cubical Ripser \cite{c:3}, built over the popular package Ripser \cite{c:11}, utilizing weighted cubical complexes to compute Betti numbers. The threshold value $\eta$ is chosen such that the first three Betti numbers (at layer-1) are non-zero for all the images in the dataset. In the case of feedforward architectures like VGG, the measure $\Omega$ is computed at each layer, while for complex architectures like ResNets/DenseNets, it is computed at each block after residual connection, considering the output volume data/tensor of shape $(w \times h \times c)$ where $w$ is the width, $h$ is the height, and $c$ is the number of channels.

\begin{figure}[ht]
    \centering
    \begin{tabular}{cc}
     \subfigure[ResNet152]{\pbox{.45\linewidth}{\includegraphics[scale=0.3]{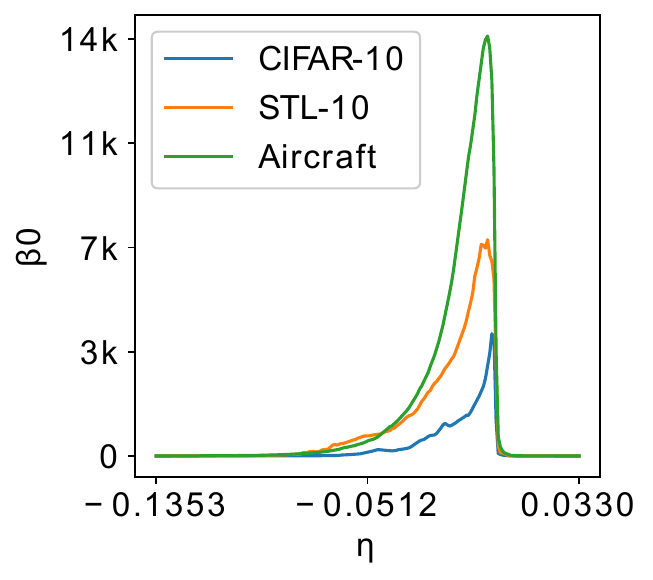} \\ \includegraphics[scale=0.3]{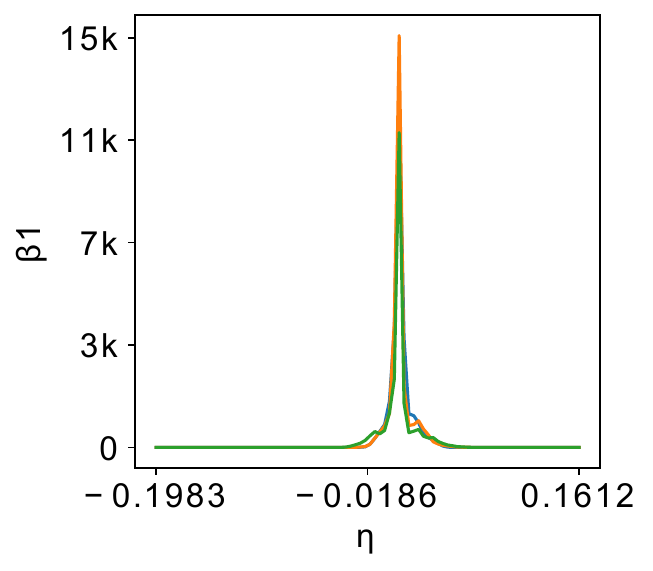} \\ \includegraphics[scale=0.3]{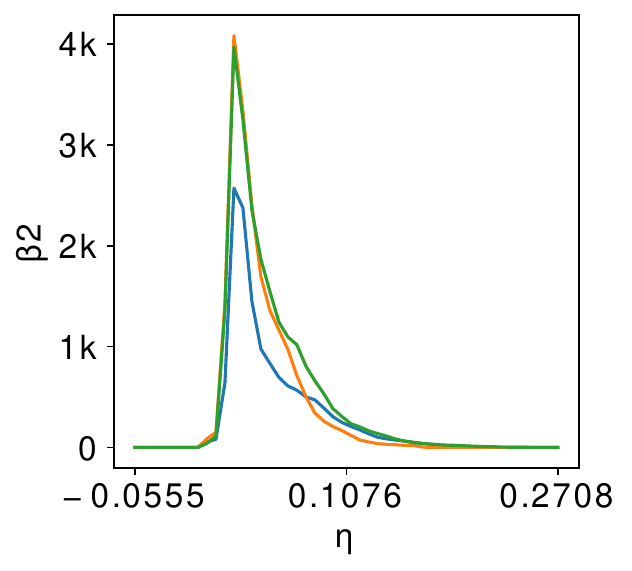}  } }    &  \subfigure[MobileNetV2]{\pbox{.55\linewidth}{\includegraphics[scale=0.3]{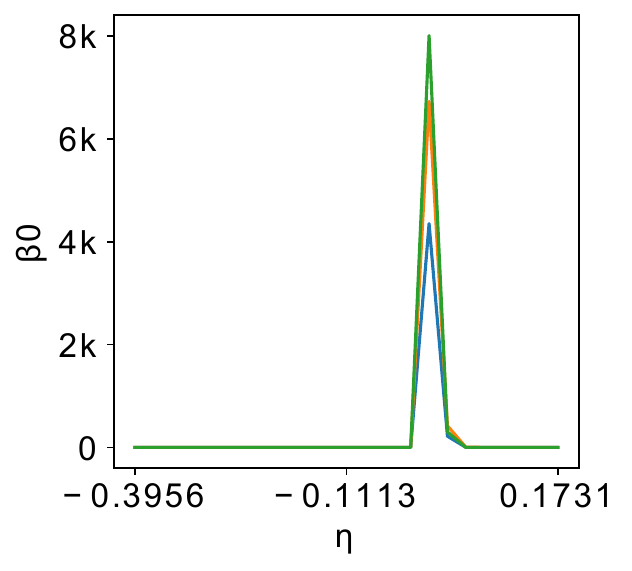} \\ \includegraphics[scale=0.3]{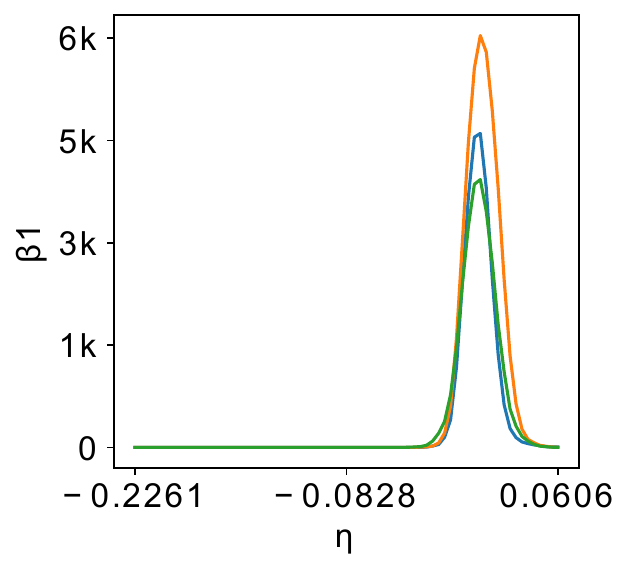} \\ \includegraphics[scale=0.3]{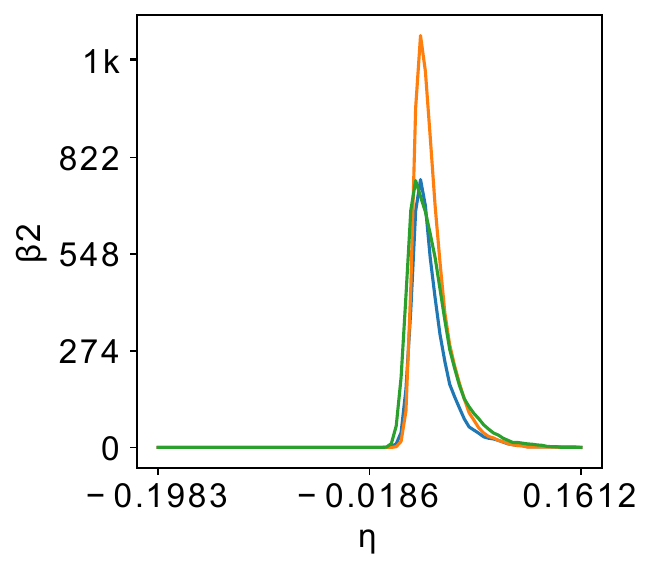} } }
    \end{tabular}
    \caption{Betti curve for $\beta_{0}$, $\beta_{1}$ and $\beta_{2}$ of the first layer embedding space of (a) ResNet-152 and (b) MobileNetV2 model, respectively.}
    \label{fig3}
\end{figure}

\section{Topology of DNN Embedding Space using Cubical Complexes}
As the inputs undergo layer-by-layer transformations in a DNN, a well-trained network should observe a decay in Betti numbers \cite{c:4}. This is because with depth, a topologically complicated dataset is transformed into a simple one, resulting in Betti numbers attaining their lowest possible value. However, since as a DNN is expected to only focus and extract relevant information; a natural question arises: \emph{if the topology induced by a pre-trained DNN is invariant to datasets?} To establish this for similar datasets consider Fig. \ref{fig3} where one can clearly observe the near invariance of the Betti curve (up to a scale) computed for the ResNet-152 model on different image datasets. This further illustrates the stability of the topological signature extracted by Betti numbers as a function of the chosen threshold.
\begin{figure}[ht]
    \centering
    \includegraphics[scale=.3
    ]{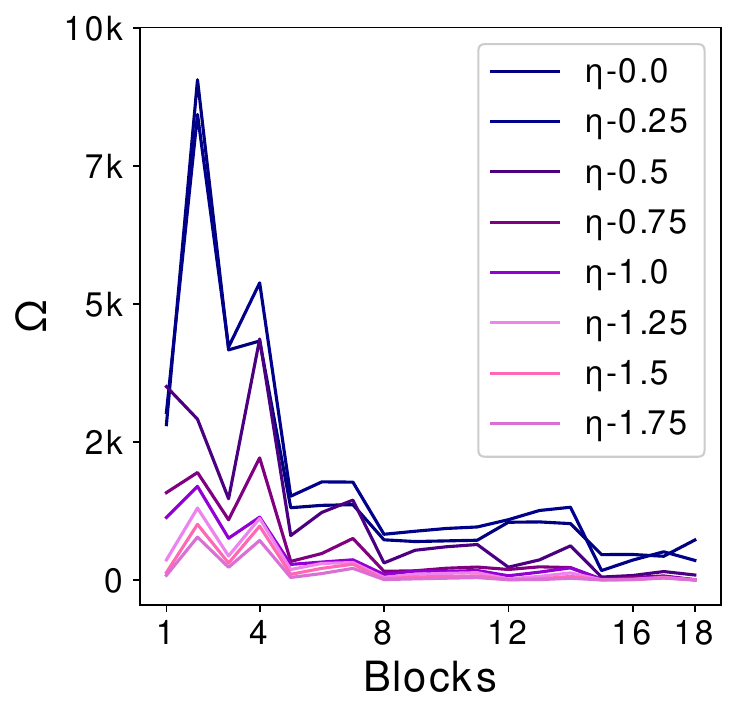}
    \caption{Topological complexity of embedding space in MobileNetV2 model for STL-10 dataset.}
    \label{fig4}
\end{figure}
\begin{figure}[ht]
    \centering
   \subfigure[VGG-16]{\includegraphics[scale=0.3]{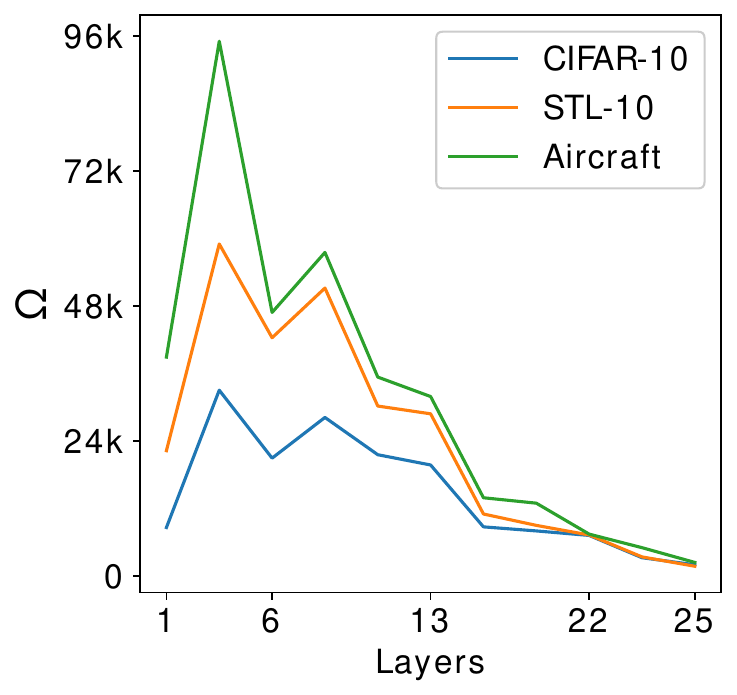}}
   \subfigure[ResNet-18]{\includegraphics[scale=0.3]{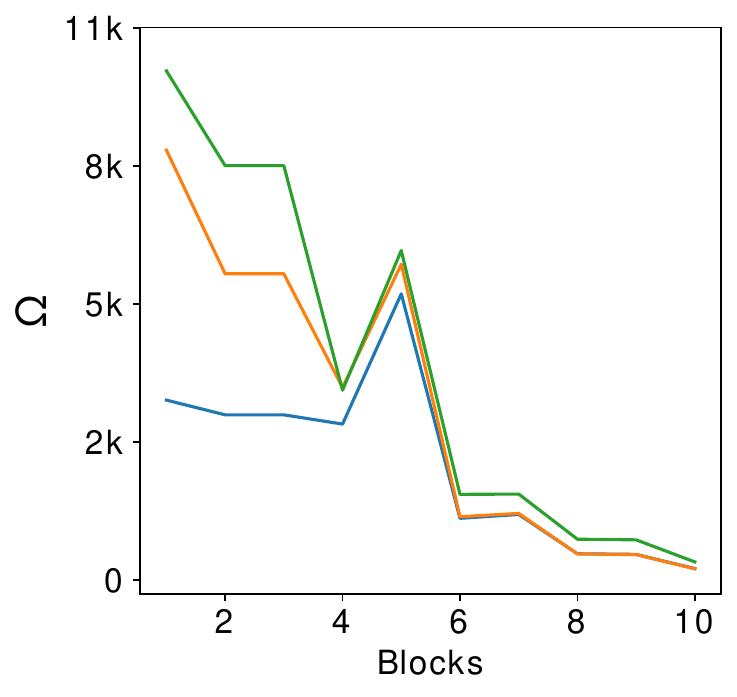}}\\
    \subfigure[MobileNetV2]{\includegraphics[scale=0.3]{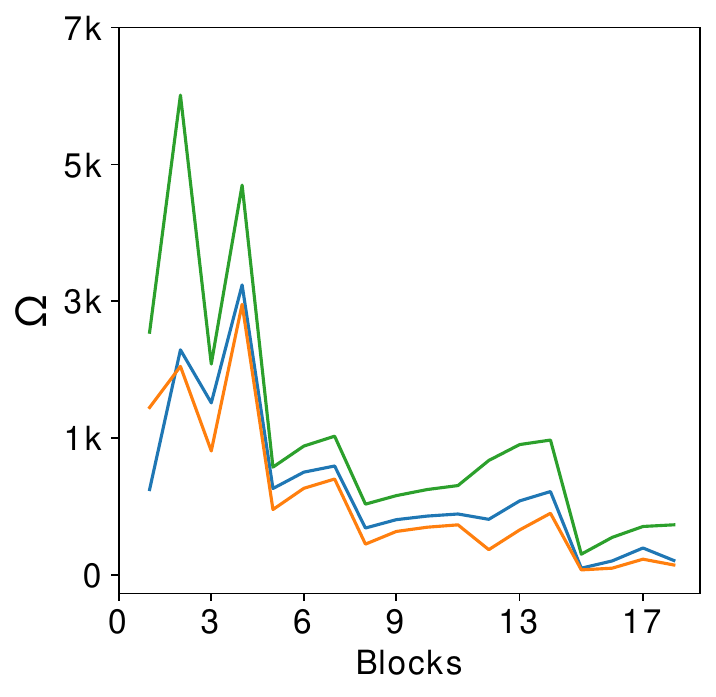}}
    \subfigure[DenseNet-121]{\includegraphics[scale=0.3]{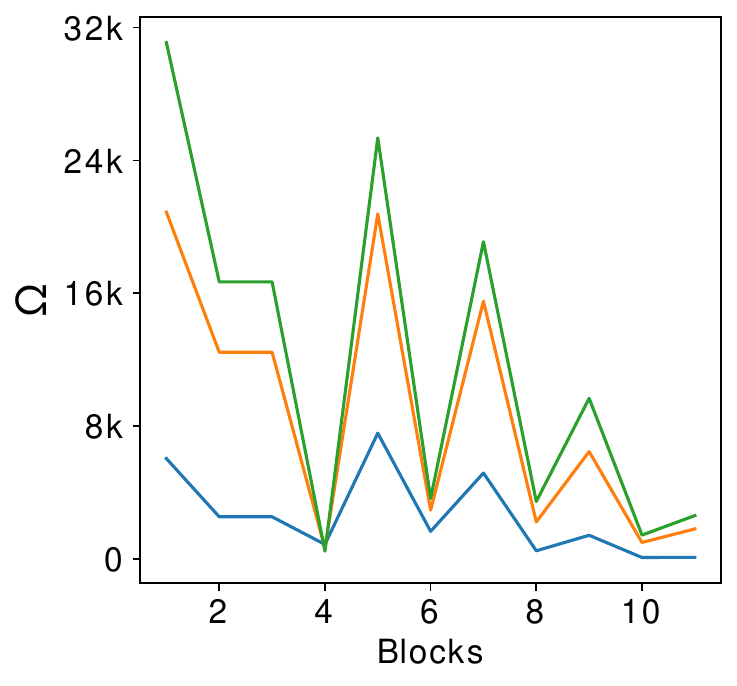}}
    \caption{Each figure show $\Omega$ values for datasets CIFAR-10,STL-10 and Aircraft. Each represents a different pattern and has the Betti numbers reduced across the layers.}
    \label{fig5}
\end{figure}

\begin{figure}[ht]
    \centering
    \includegraphics[scale=0.4]{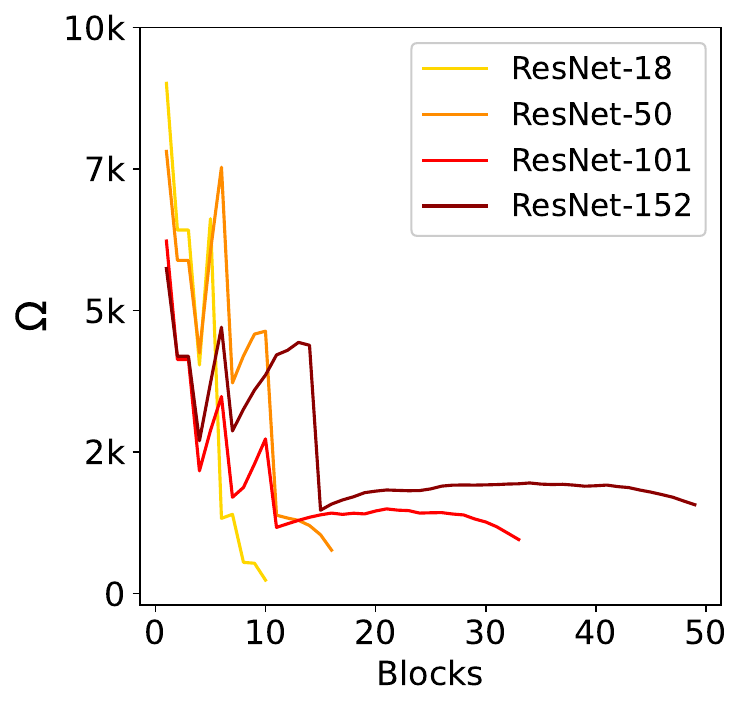}
    \caption{Topological complexity of embedding space in ResNet model of variable depth for STL-10 dataset.} 
    \label{fig6}
\end{figure}

Next, we explore the evolution of topological complexity of layer-by-layer embedding space via the proposed measure $\Omega$. Fig. \ref{fig4} shows the measure $\Omega$ as a function of the number of layers/blocks of a network. As expected due to non-homeomorphic maps (functions that don't preserve topology), we observe a decay in the value of $\Omega$ with depth. One can categorize the layers/blocks into three distinct groups of blocks: 1) the initial ones which act as low-level feature extractors, 2) the middle ones with high-level features, and 3) the penultimate ones, which can be attributed to near identity maps where $\Omega$ doesn't vary much. A notable observation is that $\Omega$ is not sensitive to the change in the value of threshold $\eta$ up to a scale that leads to robust interpretations about the learning characteristics of a DNN. One can also observe the impact of architectural choices; for instance, the number of channels directly impacts the value of $\Omega$, and hence the decay in $\Omega$ is not monotonic. As illustrated in Fig. \ref{fig5} this behavior is consistent across datasets and network architectures. In particular, for the DenseNet model, we have also considered the transient block (with a pooling layer) in between two dense blocks separately. Observe how transient block (a contraction map) impacts and drastically reduces the value of $\Omega$. Hence, one should be careful while analyzing and interpreting $\Omega$ across layers, blocks, or layers within a block, as the decay in value might not always be monotonic.

\subsection{Impact of depth on topological complexity of the embedding space of an architecture}
In the previous section, we establish the characterization of Betti numbers over depth for different architectures. However, in many tasks, the same architecture with variable depth is often compared for performance gains. In Fig. \ref{fig6} we demonstrate the behaviour of $\Omega$ for a pre-trained ResNet model of different depths. It can be observed that for a deeper architecture the number of blocks responsible for learning discriminative feature approach a limit (e.g., approx 15 for ResNet15 vs 10 for ResNet50). Also, the number of blocks for which $\Omega$ is near constant increases with depth. This limiting is interesting from a representation learning perspective, as it shows that the amount of depth required for a downstream task is limited by the modeling capability of architecture and not its capacity. This is in line with existing studies that show that while theoretically deeper architectures are more expressive but for a given dataset not all layers contribute equally \cite{c:23}. We observed similar characteristics in other architectures too, which establishes this important invariance.

\begin{figure}[t]
    \centering
    \includegraphics[scale=0.4]{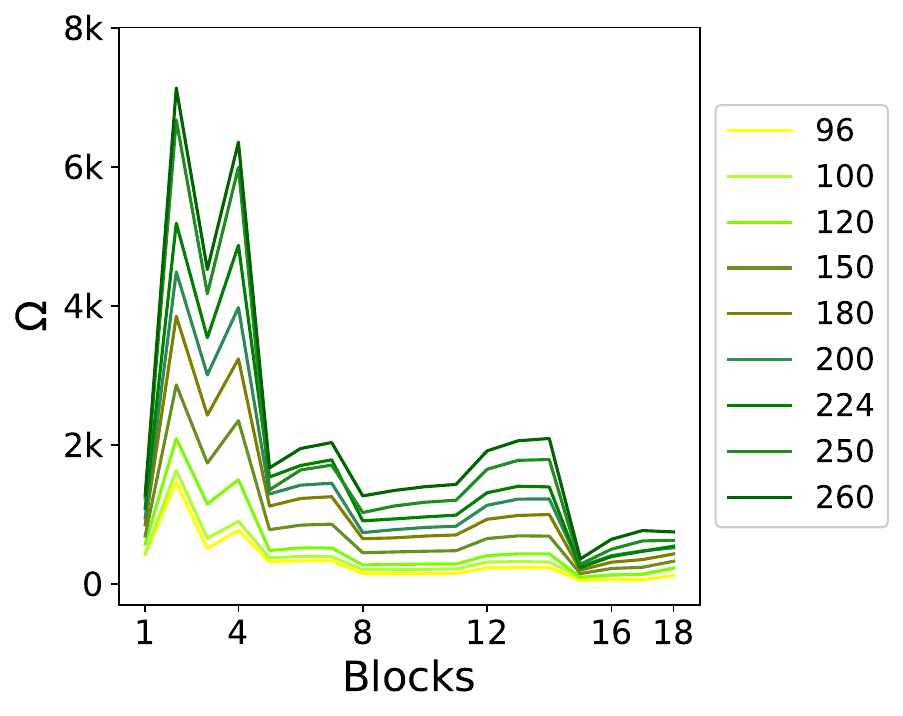}
    \caption{Topological complexity of embedding space in MobileNetV2 model for STL-10 dataset at different input resize value.} 
    \label{fig7}
\end{figure}

\subsection{Impact of input resolution on the computation of topological complexity}
In computer vision, pre-trained networks are often trained on inputs with a different resolution than the downstream task. For instance, transfer learning from the pre-trained network on the ImageNet dataset (with an image size of $256\times 256$) to the STL-10 dataset (with an image size of $96\times 96$). Hence, inputs are resized, such as by bilinear interpolation before feeding them to a pre-trained network. This makes working with topological tools on datasets with larger image sizes computationally expensive. We believe this might be the reason for earlier studies to restrict experimentation to either synthetic data or toy datasets like MNIST. Even though in the case of cubical homology, this cost increases exponentially with input size, it is near linear in the number of channels \cite{c:3}. Due to the very nature of cubical filtration in computing homology, we argue that the behaviour of $\Omega$ should be approx invariant to the input resolution. To demonstrate this, we computed the topological complexity of embedding space with inputs of variable size, and the results are reported in Fig. \ref{fig7}. It can be observed that $\Omega$ is near invariant to input size up to a scale factor. This experiment suggests that for pre-trained networks on ImageNet, input size in the range of 150 to 180 provides a good trade-off between computational complexity and explainability/interpretation of the learning dynamics of a network.

\subsection{Impact of sample complexity on the computation of topological complexity}
\begin{figure}[ht]
    \centering
    \includegraphics[scale=0.3]{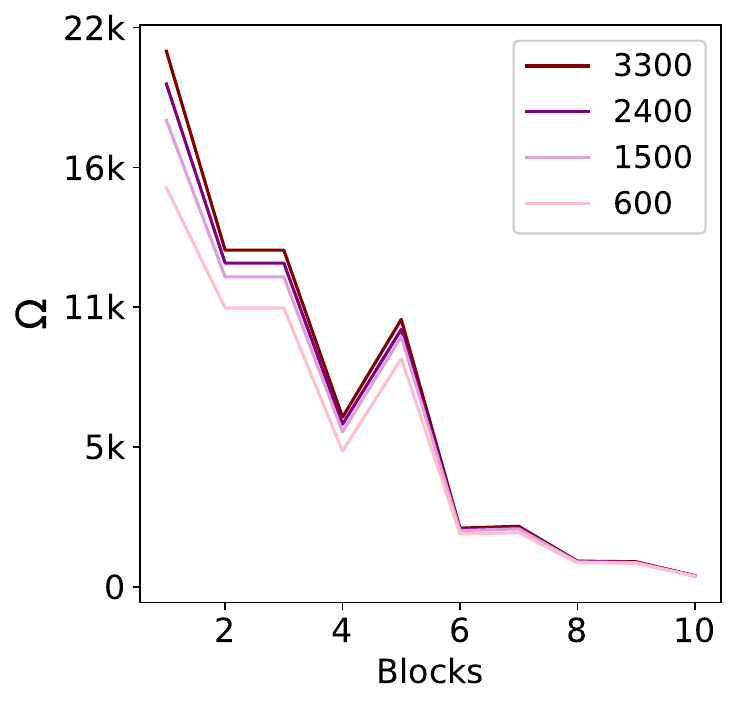}
    \caption{Topological complexity of embedding space in ResNet-18 model for Aircraft dataset with test set size ranging from 600 to 3300.} 
    \label{fig8}
\end{figure}

Similar to the input size, the categorization of embedding space using Betti numbers depends on the sample complexity \cite{c:6}. In order words, more the data points better the estimate of topological complexity. However, the computational complexity can be prohibitive for large-scale datasets. Hence, in this experiment, we explore the impact of sample complexity on the computation of the topological complexity measure. For simplicity, we consider a balanced random subset of each class, although better sampling methods can also be explored. Results in Figure \ref{fig8} demonstrate that the trajectory of $\Omega$ with depth is near invariant to sample size. This has an important implication, as the experiments with different input \& sample size suggest that a good trade-off can be achieved between computational complexity and quality of qualitative/quantitative results for DNN explainability using topology.  
% \begin{figure}[h]
%     \centering
%     \subfigure[MobileNetV2]{\includegraphics[scale=.3]{images/eps_fig/polyfit/polyfit_mobilenetv2_cifar10.eps}}
%     \subfigure[ResNet-50]{\includegraphics[scale=.3]{images/eps_fig/polyfit/polyfit_resnet50_aircraft.eps}}
%     \caption{Trajectory of $\Omega$ over depth (red) with a polynomial fit of order 2 (dotted green) and 3 (dotted blue), respectively.}
%     \label{fig9}
% \end{figure}

\begin{figure*}[h]
    \centering
    \includegraphics[trim={385 300 -2 0},clip,scale=.5]{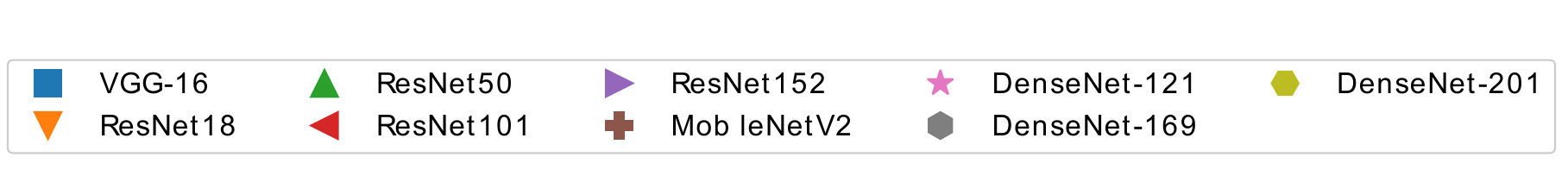}\\
    \subfigure[CIFAR-10]{\includegraphics[scale=0.4]{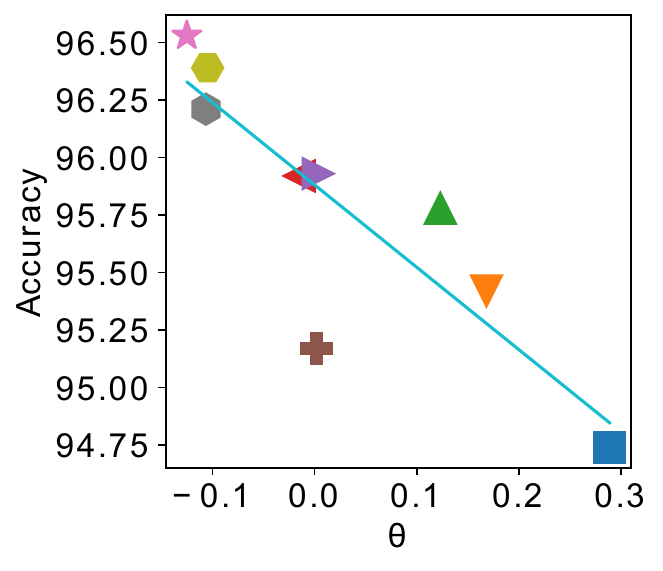}}
    \subfigure[STL-10]{\includegraphics[scale=0.4]{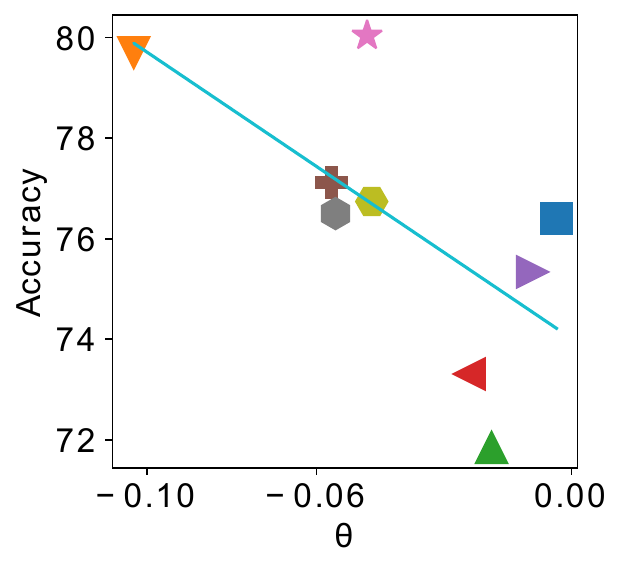}}
    \subfigure[Aircraft]{\includegraphics[scale=0.4]{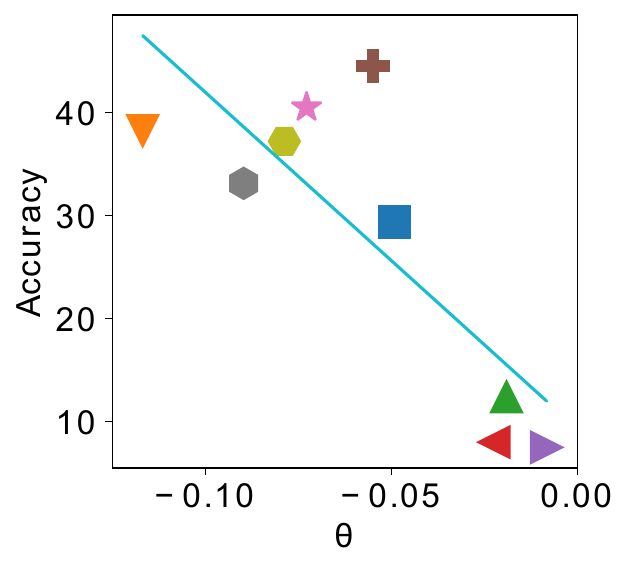}}
    \caption{Correlation between fine-tuned accuracy (Y-axis) and TTP measure ($\theta$) (X-axis) for model ranking on 3 datasets with 9 pre-trained models.}
    \label{fig10}
\end{figure*}

\begin{table}[ht]
\setlength{\extrarowheight}{.5pt}
\centering
\caption{Comparison of Pearson correlation coefficient ($\rho$) of LEEP and the proposed TTP measure. Correlations are computed with respect to test accuracies in various experimental settings}
\small
\begin{tabular}{|ccc|cc|}
\hline
\multicolumn{3}{|c|}{Experimental Setup}                                      & \multicolumn{2}{c|}{$\rho$}      \\ \hline
\multicolumn{1}{|c|}{Source} & \multicolumn{1}{c|}{Target} & Model & \multicolumn{1}{c|}{LEEP} & TTP \\ \hline
\multicolumn{1}{|c|}{\multirow{15}{*}{ImageNet}} & \multicolumn{1}{c|}{\multirow{4}{*}{STL-10}} &  ResNet-18 & \multicolumn{1}{c|}{.86} & .94 \\ \cline{3-5}
\multicolumn{1}{|c|}{} & \multicolumn{1}{c|}{} &  ResNet-50 & \multicolumn{1}{c|}{.72} & .70 \\ \cline{3-5}
\multicolumn{1}{|c|}{} & \multicolumn{1}{c|}{} &  ResNet-152 & \multicolumn{1}{c|}{.76} & .87 \\ \cline{3-5}
\multicolumn{1}{|c|}{} & \multicolumn{1}{c|}{} &  DenseNet-169 & \multicolumn{1}{c|}{.68} & .93 \\ \cline{2-5}
\multicolumn{1}{|c|}{} & \multicolumn{1}{c|}{\multirow{4}{*}{Aircraft}} &  ResNet-18 & \multicolumn{1}{c|}{.92} & .87 \\ \cline{3-5}
\multicolumn{1}{|c|}{} & \multicolumn{1}{c|}{} &  ResNet-50 & \multicolumn{1}{c|}{.82} & .89 \\ \cline{3-5}
\multicolumn{1}{|c|}{} & \multicolumn{1}{c|}{} &  DenseNet-169 & \multicolumn{1}{c|}{.68} & .81 \\ \cline{3-5}
\multicolumn{1}{|c|}{} & \multicolumn{1}{c|}{} &  DenseNet-201 & \multicolumn{1}{c|}{.78} & .92 \\ \cline{2-5}
\multicolumn{1}{|c|}{} & \multicolumn{1}{c|}{\multirow{4}{*}{CIFAR-10}} &  ResNet-50 & \multicolumn{1}{c|}{.91} & .89 \\ \cline{3-5}
\multicolumn{1}{|c|}{} & \multicolumn{1}{c|}{} &  ResNet-101 & \multicolumn{1}{c|}{.73} & .87 \\ \cline{3-5}
\multicolumn{1}{|c|}{} & \multicolumn{1}{c|}{} &  MobileNetV2 & \multicolumn{1}{c|}{.69} & .64 \\ \cline{3-5}
\multicolumn{1}{|c|}{} & \multicolumn{1}{c|}{} &  DenseNet-169 & \multicolumn{1}{c|}{.87} & .95 \\ \cline{2-5}
\multicolumn{1}{|c|}{} & \multicolumn{1}{c|}{\multirow{2}{*}{Caltech-101}} &  ResNet-50 & \multicolumn{1}{c|}{.67} & .72 \\  \cline{3-5}
\multicolumn{1}{|c|}{} & \multicolumn{1}{c|}{} &  DenseNet-169 & \multicolumn{1}{c|}{.58} & .65 \\ \hline
\end{tabular}
\label{tab1}
\end{table}

\section{Case Study: Model Ranking for Transfer Learning}
In this experiment, we establish a link between the topological complexity of a network and its generalization capability. To this aim, we consider the model ranking problem where given a target dataset and a large pool of pre-trained models, the goal is to rank the models in terms of the ease of knowledge transfer from the source model to the target dataset. This saves the computation cost involved with manually fine-tuning pre-trained models to select an optimal one. Existing studies in this space can be broadly classified into two categories with ones that compute: 1) a distribution over label space to find source and target label compatibility e.g., NCE \cite{nce}, LEEP \cite{leep}, Neural checkpoint \cite{nchk}; and 2) a distribution over feature space to find last layer embedding and target label compatibility e.g., LogME \cite{logme}, Model-Linearization \cite{linmod}. We show that one can also use the trajectory or the rate of decay of the proposed measure $\Omega$ 
to evaluate the transferability of representations learned by pre-trained classifiers. We denote this topological transferability measure (TTP) by $\theta$, computed as the instantaneous slope of the polynomial $P$ fitted to $\Omega$ values across the layers/blocks of the neural network i.e.,
\begin{equation}
    \theta = \mbox{slope} (P) |_{b},
    \label{eq2}
\end{equation}
where, $b$ is the midpoint on the curve $P$. The aim here is not to propose a state-of-the-art transferability measure but to demonstrate the power of topological tools for model interpretability. However, note that, unlike most existing measures, $\theta$ can also be applied when the source and target tasks have different label semantics or are heterogeneous.

We perform transfer learning experiments with the datasets and models described in Section~\ref{sec:exp}, where only the penultimate classification layers of the models were fine-tuned to establish the baselines. In Fig.~\ref{fig10} we show the negative correlation (inverse relationship) between the measure $\theta$ and the maximum achievable accuracy if a network is actually fine-tuned on the dataset. Table~\ref{tab1} shows the correlation coefficients of the state-of-the-art LEEP measure in comparison with the proposed measure. It can be observed that the proposed measure performs comparably to LEEP for shallow to medium-depth models. However, the proposed measure consistently performs better in the case of very deep architectures. This is because the proposed measure considers the network as a whole, while LEEP only considers the last layer. Embedding from the penultimate layer has less discrimination ability than some intermediate layers due to the input size and complexity of the target dataset; hence, the initial layer with frozen weights is often used as a feature extractor in the transfer learning framework.

\section{Conclusion}
This work attempts to demonstrate the potential of topological tools in quantifying the complexity of embedding space in a DNN. In particular, we propose a measure based on Betti numbers to quantify the topological complexity. Experiments with a variety of architectures unravel interesting invariance which provides important cues for characterizing the learning behavior in DNNs. Further,  we established a link between the expressivity \& the generalization capability of a network by considering a case study on ranking pre-trained models for quantifying the ease of transfer learning from a source to a target dataset. 

\section{Acknowledgment}
This work is supported by IIITD-IITD joint research grant (MFIRP-233/004) and Infosys Foundation via Infosys Centre for AI, IIIT Delhi.
%Bibliography
\bibliographystyle{unsrt}  
%\bibliography{references}  

\begin{thebibliography}{10}
\expandafter\ifx\csname url\endcsname\relax
  \def\url#1{\texttt{#1}}\fi
\expandafter\ifx\csname urlprefix\endcsname\relax\def\urlprefix{URL }\fi
\expandafter\ifx\csname href\endcsname\relax
  \def\href#1#2{#2} \def\path#1{#1}\fi

\bibitem{Guo21}
Y.~Guo, H.~Wang, Q.~Hu, H.~Liu, L.~Liu, M.~Bennamoun, Deep learning for 3d point clouds: A survey, IEEE Transactions on Pattern Analysis and Machine Intelligence 43~(12) (2021) 4338--4364.
\newblock \href {https://doi.org/10.1109/TPAMI.2020.3005434} {\path{doi:10.1109/TPAMI.2020.3005434}}.

\bibitem{wei22}
X.-S. Wei, Y.-Z. Song, O.~M. Aodha, J.~Wu, Y.~Peng, J.~Tang, J.~Yang, S.~Belongie, Fine-grained image analysis with deep learning: A survey, IEEE Transactions on Pattern Analysis and Machine Intelligence 44~(12) (2022) 8927--8948.
\newblock \href {https://doi.org/10.1109/TPAMI.2021.3126648} {\path{doi:10.1109/TPAMI.2021.3126648}}.

\bibitem{c:8}
B.~Poole, S.~Lahiri, M.~Raghu, J.~Sohl-Dickstein, S.~Ganguli, \href{url = {https://proceedings.neurips.cc/paper_files/paper/2016/file/148510031349642de5ca0c544f31b2ef-Paper.pdf},}{Exponential expressivity in deep neural networks through transient chaos}, in: International Conference on Neural Information Processing Systems (NeurIPS), 2016, pp. 3368--3376.
\newline\urlprefix\url{url = {https://proceedings.neurips.cc/paper_files/paper/2016/file/148510031349642de5ca0c544f31b2ef-Paper.pdf},}

\bibitem{c:6}
W.~H. Guss, R.~Salakhutdinov, On characterizing the capacity of neural networks using algebraic topology, in: NeurIPS workshop on Deep Learning: Bridging Theory and Practice, 2017, p. 1=13.
\newblock \href {https://doi.org/10.48550/arXiv.1802.04443} {\path{doi:10.48550/arXiv.1802.04443}}.

\bibitem{MURRAY22}
M.~Murray, V.~Abrol, J.~Tanner, Activation function design for deep networks: linearity and effective initialisation, Applied and Computational Harmonic Analysis 59 (2022) 117--154, special Issue on Harmonic Analysis and Machine Learning.
\newblock \href {https://doi.org/10.1016/j.acha.2021.12.010} {\path{doi:10.1016/j.acha.2021.12.010}}.

\bibitem{c:13}
F.~Chazal, B.~Michel, An introduction to topological data analysis: Fundamental and practical aspects for data scientists, Frontiers in Artificial Intelligence 4 (2021).
\newblock \href {https://doi.org/10.3389/frai.2021.667963} {\path{doi:10.3389/frai.2021.667963}}.

\bibitem{c:10}
G.~Singh, F.~Memoli, G.~Carlsson, {Topological Methods for the Analysis of High Dimensional Data Sets and 3D Object Recognition}, in: M.~Botsch, R.~Pajarola, B.~Chen, M.~Zwicker (Eds.), Eurographics Symposium on Point-Based Graphics, The Eurographics Association, 2007, pp. 91--100.
\newblock \href {https://doi.org/10.2312/SPBG/SPBG07/091-100} {\path{doi:10.2312/SPBG/SPBG07/091-100}}.

\bibitem{c:3}
S.~Kaji, T.~Sudo, K.~Ahara, Cubical ripser: Software for computing persistent homology of image and volume data (2020).
\newblock \href {https://doi.org/https://doi.org/10.48550/arXiv.2005.12692} {\path{doi:https://doi.org/10.48550/arXiv.2005.12692}}.

\bibitem{c:9}
M.~Bianchini, F.~Scarselli, On the complexity of neural network classifiers: A comparison between shallow and deep architectures, IEEE Transactions on Neural Networks and Learning Systems 25~(8) (2014) 1553--1565.
\newblock \href {https://doi.org/10.1109/TNNLS.2013.2293637} {\path{doi:10.1109/TNNLS.2013.2293637}}.

\bibitem{c:4}
G.~Naitzat, A.~Zhitnikov, L.-H. Lim, \href{http://jmlr.org/papers/v21/20-345.html}{Topology of deep neural networks}, Journal of Machine Learning Research 21~(184) (2020) 1--40.
\newline\urlprefix\url{http://jmlr.org/papers/v21/20-345.html}

\bibitem{c:20}
G.~E. Carlsson, T.~Ishkhanov, V.~de~Silva, A.~Zomorodian, On the local behavior of spaces of natural images, International Journal of Computer Vision 76~(1) (2008) 1--12.
\newblock \href {https://doi.org/10.1007/s11263-007-0056-x} {\path{doi:10.1007/s11263-007-0056-x}}.

\bibitem{c:17}
B.~Rieck, M.~Togninalli, C.~Bock, M.~Moor, M.~Horn, T.~Gumbsch, K.~Borgwardt, \href{https://openreview.net/forum?id=ByxkijC5FQ}{Neural persistence: {A} complexity measure for deep neural networks using algebraic topology}, in: International Conference on Learning Representations~(ICLR), 2019, pp. 1--11.
\newline\urlprefix\url{https://openreview.net/forum?id=ByxkijC5FQ}

\bibitem{c:16}
M.~Bianchini, F.~Scarselli, On the complexity of neural network classifiers: A comparison between shallow and deep architectures, IEEE Transactions on Neural Networks and Learning Systems 25~(8) (2014) 1553--1565.
\newblock \href {https://doi.org/10.1109/TNNLS.2013.2293637} {\path{doi:10.1109/TNNLS.2013.2293637}}.

\bibitem{c:18}
N.~Hamada, K.~Goto, Data-driven analysis of pareto set topology, in: Genetic and Evolutionary Computation Conference, 2018, p. 657–664.
\newblock \href {https://doi.org/10.1145/3205455.3205613} {\path{doi:10.1145/3205455.3205613}}.

\bibitem{c:19}
N.~Akai, T.~Hirayama, H.~Murase, Experimental stability analysis of neural networks in classification problems with confidence sets for persistence diagrams, Neural Networks 143 (2021) 42--51.
\newblock \href {https://doi.org/https://doi.org/10.1016/j.neunet.2021.05.007} {\path{doi:https://doi.org/10.1016/j.neunet.2021.05.007}}.

\bibitem{c:1}
T.~K. Dey, Y.~Wang, Computational Topology for Data Analysis, Cambridge University Press, 2022.
\newblock \href {https://doi.org/10.1017/9781009099950} {\path{doi:10.1017/9781009099950}}.

\bibitem{c:12}
S.~Choe, S.~Ramanna, Cubical homology-based machine learning: An application in image classification, MDI Axioms 11~(3) (Mar 2022).
\newblock \href {https://doi.org/10.3390/axioms11030112} {\path{doi:10.3390/axioms11030112}}.

\bibitem{c:22}
X.~Hu, F.~Li, D.~Samaras, C.~Chen, \href{https://proceedings.neurips.cc/paper_files/paper/2019/file/2d95666e2649fcfc6e3af75e09f5adb9-Paper.pdf}{Topology-preserving deep image segmentation}, in: H.~Wallach, H.~Larochelle, A.~Beygelzimer, F.~d\textquotesingle Alch\'{e}-Buc, E.~Fox, R.~Garnett (Eds.), Advances in Neural Information Processing Systems, Vol.~32, Curran Associates, Inc., 2019, pp. 1--12.
\newline\urlprefix\url{https://proceedings.neurips.cc/paper_files/paper/2019/file/2d95666e2649fcfc6e3af75e09f5adb9-Paper.pdf}

\bibitem{c:21}
J.~W. Milnor, Morse theory, no.~51, Princeton university press, 1963.

\bibitem{coates2011stl10}
A.~Coates, A.~Ng, H.~Lee, \href{https://proceedings.mlr.press/v15/coates11a.html}{An analysis of single-layer networks in unsupervised feature learning}, in: G.~Gordon, D.~Dunson, M.~Dudík (Eds.), Proceedings of the Fourteenth International Conference on Artificial Intelligence and Statistics (AISTATS), Vol.~15 of Proceedings of Machine Learning Research, PMLR, Fort Lauderdale, FL, USA, 2011, pp. 215--223.
\newline\urlprefix\url{https://proceedings.mlr.press/v15/coates11a.html}

\bibitem{Krizhevsky09learningmultiple}
A.~Krizhevsky, \href{https://www.cs.toronto.edu/~kriz/cifar.html}{Learning multiple layers of features from tiny images}, Tech. rep., University of Toronto (2009).
\newline\urlprefix\url{https://www.cs.toronto.edu/~kriz/cifar.html}

\bibitem{maji13fine-grained}
S.~Maji, J.~Kannala, E.~Rahtu, M.~Blaschko, A.~Vedaldi, \href{https://www.robots.ox.ac.uk/~vgg/data/fgvc-aircraft/}{Fine-grained visual classification of aircraft}, Tech. rep., University of Oxford (2013).
\newblock \href {http://arxiv.org/abs/1306.5151} {\path{arXiv:1306.5151}}.
\newline\urlprefix\url{https://www.robots.ox.ac.uk/~vgg/data/fgvc-aircraft/}

\bibitem{simonyan2014very}
K.~Simonyan, A.~Zisserman, Very deep convolutional networks for large-scale image recognition, in: Y.~Bengio, Y.~LeCun (Eds.), International Conference on Learning Representations, {ICLR}, 2015, pp. 1--14.
\newblock \href {https://doi.org/10.48550/arXiv.1409.1556} {\path{doi:10.48550/arXiv.1409.1556}}.

\bibitem{he2016deep}
K.~He, X.~Zhang, S.~Ren, J.~Sun, Deep residual learning for image recognition, in: IEEE Conference on Computer Vision and Pattern Recognition (CVPR), 2016, pp. 770--778.
\newblock \href {https://doi.org/10.1109/CVPR.2016.90} {\path{doi:10.1109/CVPR.2016.90}}.

\bibitem{huang2017densely}
G.~Huang, Z.~Liu, L.~Van Der~Maaten, K.~Q. Weinberger, Densely connected convolutional networks, in: IEEE Conference on Computer Vision and Pattern Recognition (CVPR), 2017, pp. 2261--2269.
\newblock \href {https://doi.org/10.1109/CVPR.2017.243} {\path{doi:10.1109/CVPR.2017.243}}.

\bibitem{sandler2018mobilenetv2}
M.~Sandler, A.~Howard, M.~Zhu, A.~Zhmoginov, L.-C. Chen, Mobilenetv2: Inverted residuals and linear bottlenecks, in: IEEE/CVF Conference on Computer Vision and Pattern Recognition, 2018, pp. 4510--4520.
\newblock \href {https://doi.org/10.1109/CVPR.2018.00474} {\path{doi:10.1109/CVPR.2018.00474}}.

\bibitem{ILSVRC15}
O.~Russakovsky, J.~Deng, H.~Su, J.~Krause, S.~Satheesh, S.~Ma, Z.~Huang, A.~Karpathy, A.~Khosla, M.~Bernstein, A.~C. Berg, L.~Fei-Fei, {ImageNet Large Scale Visual Recognition Challenge}, International Journal of Computer Vision (IJCV) 115~(3) (2015) 211--252.
\newblock \href {https://doi.org/10.1007/s11263-015-0816-y} {\path{doi:10.1007/s11263-015-0816-y}}.

\bibitem{c:11}
U.~Bauer, Ripser: efficient computation of vietoris-rips persistence barcodes, Springer Journal of Applied and Computational Topology (2021).
\newblock \href {https://doi.org/10.1007/s41468-021-00071-5} {\path{doi:10.1007/s41468-021-00071-5}}.

\bibitem{c:23}
C.~Zhang, S.~Bengio, Y.~Singer, \href{http://jmlr.org/papers/v23/20-069.html}{Are all layers created equal?}, Journal of Machine Learning Research 23~(67) (2022) 1--28.
\newline\urlprefix\url{http://jmlr.org/papers/v23/20-069.html}

\bibitem{nce}
A.~Tran, C.~Nguyen, T.~Hassner, Transferability and hardness of supervised classification tasks, in: IEEE/CVF International Conference on Computer Vision (ICCV), 2019, pp. 1395--1405.
\newblock \href {https://doi.org/10.1109/ICCV.2019.00148} {\path{doi:10.1109/ICCV.2019.00148}}.

\bibitem{leep}
C.~Nguyen, T.~Hassner, M.~Seeger, C.~Archambeau, \href{https://proceedings.mlr.press/v119/nguyen20b.html}{{LEEP}: A new measure to evaluate transferability of learned representations}, in: H.~D. III, A.~Singh (Eds.), International Conference on Machine Learning (ICML), Vol. 119 of Proceedings of Machine Learning Research, PMLR, 2020, pp. 7294--7305.
\newline\urlprefix\url{https://proceedings.mlr.press/v119/nguyen20b.html}

\bibitem{nchk}
Y.~Li, X.~Jia, R.~Sang, Y.~Zhu, B.~Green, L.~Wang, B.~Gong, Ranking neural checkpoints, in: IEEE/CVF Conference on Computer Vision and Pattern Recognition (CVPR), 2021, pp. 2662--2672.
\newblock \href {https://doi.org/10.1109/CVPR46437.2021.00269} {\path{doi:10.1109/CVPR46437.2021.00269}}.

\bibitem{logme}
K.~You, Y.~Liu, J.~Wang, M.~Long, Logme: Practical assessment of pre-trained models for transfer learning, in: International Conference on Machine Learning (ICML), 2021, pp. 12133--12143.

\bibitem{linmod}
A.~Deshpande, A.~Achille, A.~Ravichandran, H.~Li, L.~Z. andCharless Fowlkes, R.~Bhotika, S.~Soatto, P.~Perona, \href{https://arxiv.org/abs/2102.00084}{A linearized framework and a new benchmark for model selection for fine-tuning} (2021).
\newline\urlprefix\url{https://arxiv.org/abs/2102.00084}

\end{thebibliography}

\end{document}